# Will People Like Your Image? Learning the Aesthetic Space

Katharina Schwarz   Patrick Wieschollek   Hendrik P. A. Lensch
University of Tübingen

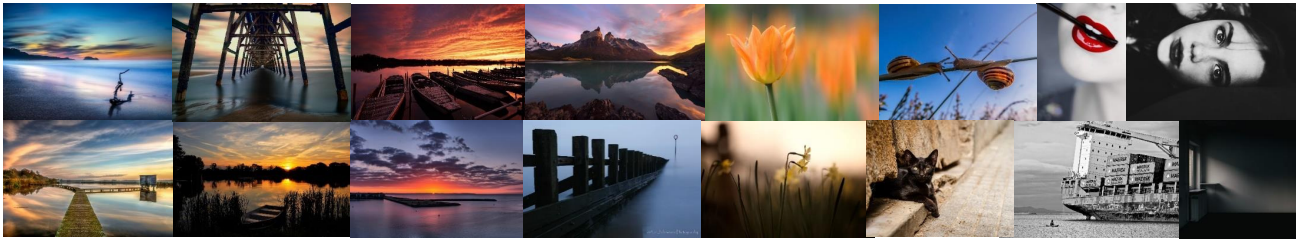

Figure 1. Aesthetically pleasing images from our derived scores. The complex matter of aesthetics depends on many factors, e.g., visual appearance, composition, content, or style, and makes it almost impossible to directly compare all images if they are similarly beautiful.

## Abstract

*Rating how aesthetically pleasing an image appears is a highly complex matter and depends on a large number of different visual factors. Previous work has tackled the aesthetic rating problem by ranking on a 1-dimensional rating scale, e.g., incorporating handcrafted attributes. In this paper, we propose a rather general approach to automatically map aesthetic pleasingness with all its complexity into an "aesthetic space" to allow for a highly fine-grained resolution. In detail, making use of deep learning, our method directly learns an encoding of a given image into this high-dimensional feature space resembling visual aesthetics.*

*Additionally to the mentioned visual factors, differences in personal judgments have a large impact on the likeableness of a photograph. Nowadays, online platforms allow users to "like" or favor certain content with a single click. To incorporate a huge diversity of people, we make use of such multi-user agreements and assemble a large data set of 380K images (AROD) with associated meta information and derive a score to rate how visually pleasing a given photo is. We validate our derived model of aesthetics in a user study. Further, without any extra data labeling or handcrafted features, we achieve state-of-the art accuracy on the AVA benchmark data set. Finally, as our approach is able to predict the aesthetic quality of any arbitrary image or video, we demonstrate our results on applications for resorting photo collections, capturing the best shot on mobile devices and aesthetic key-frame extraction from videos.*

## 1. Introduction

The wide distribution of digital devices allows us to take series of photos making sure not to miss any big moment. Manually picking the best shots afterwards is not only time-consuming but also challenging. Generally, approaches for automatically ranking images towards their aesthetic appeal can be useful in many applications, e.g., to handle personal collections or for retrieval tasks. Overall, deciding how aesthetically pleasing an image appears is always a highly complex matter depending on a large number of various factors: Visual appearance, image composition, displayed content, or style all influence its aesthetic appeal. Fig. 1 shows a set of beautiful images with different appearance. Assume one would score each of them separately, e.g., within the interval $[1, 10]$ to obtain some granularity. This is not only a challenging task. Even more critical, mapping those ratings to an *absolute* scale afterwards can lead to wrong relations between them. Asking for *relative* comparisons is not only an easier task, but also leads to a more reliable scale. For images like in Fig. 1 it is still almost impossible to directly compare all of them, e.g., the beautiful warmth of a sunset can hardly be generally related to the coolness of an image in style "noir". Overall, it is often unclear which particular attribute mostly influences the aesthetic comparison of an image pair. Thus, we propose to arrange images in a high-dimensional space to obtain a better understanding on a highly fine-granular level about how the aesthetic appeal correlates between them without predefining specific factors. On saliency maps, we further demonstrate the necessity of considering global features in aesthetic tasks.

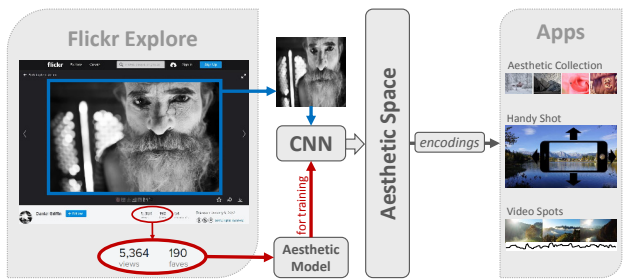

Figure 2. Overview. Based on images we assemble from Flickr, we derive a model that scores aesthetic appeal of an image from its "views" and "faves". This model then guides the training process to learn fine-grained relations in the high-dimensional "aesthetic space". Finally, our trained CNN is able to generate encodings for any arbitrary image leading to several applications.

Additionally to the previously mentioned factors, differences in personal judgments have a large impact on the likeableness of a photograph. Nowadays, online platforms allow users to "like" or favor certain content with a single click. Usually people "like" beautiful images or, in other words, aesthetically pleasing ones. Sometimes, people might also favor images for other reasons like based on their scene content, e.g., picturing the newest mobile phone. Anyway, our user study shows that our derived model is still reliable. In this work, we consider both, the complexity of aesthetics in its high-dimensionality as well as a huge diversity of multi-user online ratings to obtain broad information about aesthetic relations without extra data labeling.

An overview of our method is illustrated in Fig. 2. First, we assemble a large amount of images from Flickr and present a new database to exploit *Aesthetic Ratings from Online Data* (AROD). Therefrom, we derive a model of aesthetics to score the quality of an image by making use of the huge amount of available online user behavior, the "views" and "faves". Then, we make use of deep learning and include the introduced measurements of aesthetic appeal indirectly as hints to guide the training process. Thereby, we only incorporate the information if two images are aesthetically similar or not instead of using the direct score. This allows us to consider every single image relatively to other images – even if they do not seem visually comparable, i.e., due to large differences in their visual factors like appearance, displayed content, or style. Our trained CNN is then able to directly learn an encoding of any given image in a high-dimensional feature space resembling visual aesthetics. Our "aesthetic space" encodes the complex matter of aesthetics, that not every pair can be directly compared, on a highly fine-grained resolution of relative distances. Finally, as those encodings can be obtained for any arbitrary image, we demonstrate how they can be easily transferred into several applications on images as well as videos. In summary, our main contributions are:

- A new large-scale data set containing dense and diverse meta information and statistics to reliably predict visual aesthetics and which is easily extendable.
- A model that approximates aesthetic ratings on a broad diversity without specifically requesting expensive labels beforehand and which we validate in a user study.
- Formulating the complexity of aesthetic prediction as an encoding problem to directly learn the feature space allowing for fine-granularity of relative rankings on a high-dimensional level.
- Application prototypes such as an app for mobile devices, a photo-collection manager powered by visual aesthetic prediction as well as a video processing tool to score frames.

## 2. Related Work

**Aesthetics in Images.** Previous research on visual aesthetics assessment focused on handcrafted visual cues such as color [37, 6, 36], texture [6, 19], or content [7, 30].

Generally, no absolute rules exist to ensure high aesthetic quality of a photograph. Photo quality has been explored to distinguish between high and low quality [19] or classify between the aesthetic quality of a photograph taken from a professional vs a laymen [40]. Besides of quality, interest has arisen towards the importance of images. Thereby, previous work has exploited if and to which extent an image can be predicted as "popular" [20], "memorable" [14], or "interesting" [7, 11, 8]. Thereby, aesthetics played roles like how it influences the memorability of an image [14]. Also, the relationship between aesthetics and images has been explored from multiple perspectives [17]. Further, making use of deep learning, the "style" of an image has been of recent interest: either to recognize a specific image style [18] or even to manipulate images by transferring artistic style from a painting to a captured photo [9, 16]. Such style attributes have been incorporated to improve aesthetic categorization [28]. In addition to style, the composition of an image largely influences aesthetic pleasingness and has been explored in terms of rules or enhancement [15, 26, 10]. Overall, many approaches have investigated a lot of work to find adequate attributes to approach aesthetics, e.g., generic image descriptors [32], attributes humans might use [7], cues performing psychological experiments [11], features based on artistic intuition [6], content-based features [30], or features with high computational efficiency [27]. Other methods have focused on classifying the aesthetic appeal restricting their content to consumer photos with faces [23, 24], consumer videos [34, 1] or other visual domains, e.g., paintings [22] or evolved abstract images [4]. In contrast to those previous methods, we aim for a general approach to explore the global overall aesthetic appeal without any necessity to restrict image content or define any specific attributes or properties.

Table 1. Comparison of different data sets containing images for judging visual pleasingness of images. * Per image

| properties | AVA [35] | AADB [21] | **AROD (ours)** |
|---|---|---|---|
| max ratings* | 549 | 5 | **2.8M** |
| mean ratings* | 210 | 5 | **6868** |
| rating distr. | normal | normal | **uniform** |
| number of images | 250K | 10K | **380K** |
| avg. image size | 602×689 | 773×955 | **1926×2344** |

**Deep Metric Learning.** Neural networks are capable of organizing arbitrary input in a latent space. Approaches directly manipulating this space have been successfully applied to signature verification [2], face recognition [5, 41] and comparing image patches [42] for depth estimation. Hereby, feature representations of the inputs are optimized such that they describe similarity relations within the data. Therefore, metric learning methods such as Siamese networks [5] and Triplet networks [13] are widely used. Inspired by those successful networks, we now approach the aesthetic learning problem by directly optimizing a metric to position aesthetic relations in a high-dimensional space.

**Deep Learning Aesthetics.** Transferring aesthetics into a deep learning approach without defining hand-crafted features has been formulated as a categorization problem based on extracting patches for training [28, 29]. However, reducing visual content to small patches can destroy the global appearance which is important for aesthetic tasks. In contrast, we incorporate the entire image and demonstrate the importance of global features on saliency maps.

Other methods have considered image quality rating as a traditional classification or regression problem predicting a single scalar information real or binary [35, 21]. Thus, they do not meet the complex nature of aesthetics as they oversimplify the task. They focus on a single scale problem that even humans might not be able to solve as they probably disagree on the actual level of visual pleasingness. Further, these approaches either use hand-crafted features [21] or examine a data set of small annotation density [35, 21]. In contrast to those methods, we make use of deep metric learning to transfer the problem of aesthetic ranking into a high-dimensional feature space representation. We rely on the plain image without defining any kinds of attributes.

## 3. Data Sets

Generally, the training of deep networks requires large annotated data sets [38, 25] to obtain reliable results. Further, as visual aesthetics of photos is highly subjective depending on the current mood as well as any emotion, training a data-driven model requires extensive, diverse annotations. To overcome flaws of previous benchmark sets, we introduce a new data set with a comparison given in Table 1.

### 3.1. Previous Data Sets

**AVA.** The AVA data set [35] provides 250K images classified in visually well-crafted and mediocre ones on a fix scale. These images are obtained from a professional community of photographic challenges. Through their annotation process only a very small amount of annotations are collected in comparison to the dimensions of social network members comprising also non-professional photographers. Note, to reliably judge image aesthetics it is inevitable to consider the consensus of highly diverse participants.

**AADB.** Recently, Kong *et al*. [21] introduced a new aesthetics and attributes data set (AADB) comprising of 10K images. Each individual image score in AADB represents the averaged rating of five AMT (Amazon Mechanical Turk) workers, who are *asked* to give each image an overall aesthetic score. In addition, they provide attribute assignments from 11 predefined categories as judged by AMT workers. Their database maintains photos downloaded directly from Flickr which are likely to be not post-processed in contrast to professional results contained in AVA [35].

### 3.2. Our Flickr Subset

Whereas AADB is quite small, the image data of AVA seems rather biased. Besides, both only provide a small amount of collected ratings (Table 1). Thus, we propose a new, much larger data set comprising aesthetic ratings from online data (AROD). This data can be downloaded immediately, including meta-data as well as extensive, diverse labels, without the need to collect extra ratings spending additional time, effort, and money.

**AROD.** A single click allows users to give feedback to media content. We propose to use this information. E.g., Flickr allows to add any photo to a personal list of favorites, which is counted as "faves". Since this feature is optionally, users are absolutely free to add a particular image to their favorite list. Their only motivation is to tag a photo which is worth to remember. In addition, these images are uploaded without a purpose to participate in a concrete challenge and are not limited to a specific topic.

To collect these image we crawl around 380K photos from Flickr including meta data such as their number of views, comments, favorite list containing this photo, title of the image and their description from the Flickr website. Our collection contains images which were published and uploaded between January 2004 and November 2016. As each photo is visited ∼7K times in average, this allows for a much finer granularity and gives more hints about aesthetics of images compared to previous data sets. Based on this data, we derive a model to obtain information about aesthetic pleasingness of the underlying image.

## 4. Model of Aesthetics

In online platforms, people usually tend to "like" beautiful images or, in other words, aesthetically pleasing ones. Thus, we now aim to explore those multi-user agreements and turn them into a new useful measurement towards aesthetic appeal. We extract time-independent statistics, the "faves" and "views" (Fig. 2), which contain information traits about the underlying image quality.

### 4.1. Model Definition

Previous attempts tried to directly regress some score or trained a simple binary model [36, 6] to decide whether an image is visually pleasing or ordinary. To overcome the classification approaches Kong *et al.* [21] employ a modification of the Siamese loss-function [2] to re-rank images according their predicted aesthetic score. In contrast to [21, 36, 6], we will leverage traits from freely available information in social networks to score the image quality. These statistics are only used as hints to guide the training process rather than as a direct label or score.

To judge the pleasingness of an image we examine the relation between the *"views"* (number of visits) and the *"faves"* (number of clicks that favor image) as a proxy for visual aesthetics. Both these landmarks are highly dependent of visual aesthetics and encode the visual quality in all its facets. In addition, the low hurdle of creating a feedback ("like" or "favor") allows to average information being orders of magnitude larger compared to data sets obtained via AMT. This is especially necessary, when treating images which are highly debatable. As common in population dynamics we assume exponential increase of the views $\frac{dV(i)}{dt} = r_{V(i)} \cdot V(i)$ and the faves $\frac{dF(i)}{dt} = r_{F(i)} \cdot F(i)$ over time $t \in \mathbb{N}$ for any arbitrary image $i \in \mathcal{I}$ with growth rate $r_{(\cdot)} > 0$. This allows us to approximate the score $S(i)$ of the image quality –independent of time $t$– by

$$S(i) \sim \frac{\log F(i)}{\log V(i)}. \qquad (1)$$

This time-independence of any image $i$ is necessary when using images with different online life-spans. In addition, the model in Eq. (1) accounts for the effect of getting more faves per image being a popular user at Flickr due to the mechanism of followers. Note, the action not to add an image to ones "faves" contains valuable information, too! Considering the score $S(i)$ gives a criteria to rank images $i \in \mathcal{I}$, which values can be imitated by neural networks (see Fig. 3). A histogram of the distribution of $S(i)$ (Eq. (1)) is illustrated in Fig. 4. The uniform distribution of the data shows that the data has high entropy which allows us to even judge borderline images.

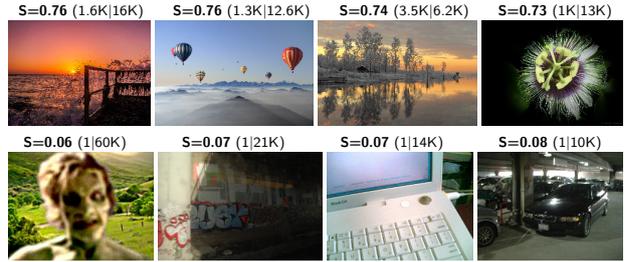

Figure 3. Images from our data set with approximated score $S$ from (#faves|#views). The upper rows shows images $i$ with large values in $S(i)$ and the bottom row with relatively low scores $S(i)$.

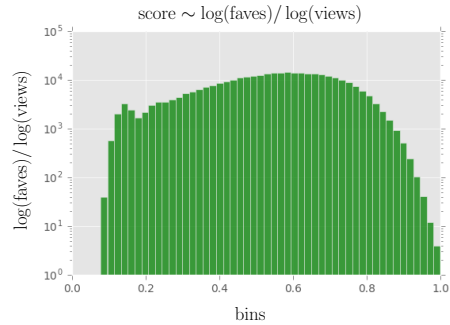

Figure 4. Distribution of the collected score $S(i)$. The uniform distribution allows us to even judge images with borderline ratings.

### 4.2. Human Evaluation

As we introduce our aesthetic model as a score based on online behavior from uncontrolled user clicks, we validate the usefulness of our derived metric in a controlled experiment. We formulate our hypotheses $\mathcal{H}$ as follows:

$\mathcal{H}_1$: Our derived "aesthetic model" based on freely available ratings from an uncontrolled human online behavior is reasonable.

$\mathcal{H}_2$: Higher scored images are also rated better in a controlled user study and worse ones are also rated worse.

Rating the aesthetic quality of an image is highly subjective and differs between persons. Performing a user study over a diversified crowd is inevitable to validate trends. As stated by Buhrmester *et al.* [3], Amazon Mechanical Turk (AMT) yields reliable data on a demographically diverse level. Thus, we use AMT to evaluate our aesthetic model.

**Experiment Setup.** To overcome differences in internal ratings between persons, we aim for *relative ratings* instead of an absolute scale. Further, to ensure that images obtaining a higher score are really more pleasant than lower scored ones, we design the study as pairwise preference tests. The AMT workers are presented two images with different scores as shown in Fig. 5. In each binary forced-

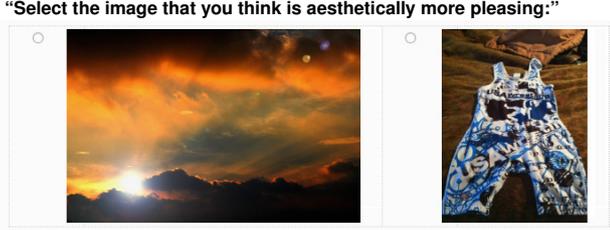

"Select the image that you think is aesthetically more pleasing:"

Figure 5. Example as seen by AMT workers. The task (top) is to select one image of the presented pair (bottom).

choice task, the Turker is asked to select the image that is "aesthetically more pleasing". We directly ask for aesthetic selection to ensure that our score derived from online "faves" is a suitable measure to rate aesthetics. From our downloaded data set, we evaluate 700 randomly selected image pairs. Each pair is presented to 5 Turkers. To negate click biases, ordering as well as positioning are randomized.

**User Study Results.** In our user study, we randomly test image pairs with varying distances between the scores derived by our model. Thereby, the lowest scored images obtained at least one fave. All evaluated distances are listed in Table 2. Thereby, a small distance means that our derived scores are very similar and that the images are almost identically pleasing towards aesthetics. However, setting the minimal distance between the scores of the 2 images in a pair to 0.1 is rated towards the similar direction by already 78% of the Turkers. Further, for score distances bigger than 0.4, even 89% of the test persons agreed with the selection of the better image. Overall, we obtain ratings with surprisingly small variance. Besides, the already relatively small variance even further decreases with increasing distance. This indicates a high agreement between the different Turkers. As verified with a Kolmogorov-Smirnov test [33], the underlying data does not come from a normal distribution. Thus, we verified statistical relevance performing the Mann-Whitney U-test [31] which rejected the null hypothesis for all distances at least at the 10% level ($p < 0.10$) and for $\Delta > 0.3$ at the 5% level ($p < 0.05$) revealing statistical significant dependency between the scores of our model and the user study ratings ($\mathcal{H}_2$). As we explicitly ask the Turkers to rate due to the term "aesthetically pleasing", our presented score $S(i)$ can really be seen as an aesthetic measure validating our first hypothesis $\mathcal{H}_1$.

Table 2. User study results. More similar rating decisions of Turkers are obtained for larger distances $\Delta = |S(i) - S(j)|$ between our derived scores $S(\cdot)$ of the images within a pair.

| dist $\Delta$ | $> 0.1$ | $> 0.2$ | $> 0.3$ | $> 0.4$ | $> 0.5$ | $> 0.6$ |
|---|---|---|---|---|---|---|
| mean $\mu$ | 0.78 | 0.85 | 0.88 | 0.89 | 0.89 | 0.89 |
| var $\sigma^2$ | 0.07 | 0.05 | 0.04 | 0.04 | 0.04 | 0.04 |
| sign. level $\alpha$ | 10% ($p < 0.10$) | | 5% ($p < 0.05$) | | | |

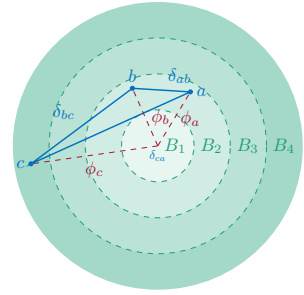

Figure 6. Previous approaches treat aesthetic learning as a low-dimensional problem [21] which projects encodings on a 1-dimensional or into discrete bins [35]. Rather than learning a bin-mapping for each image $i \in \{a, b, c\}$ into bins $B_i$ or directly $\phi_i$, we propose to learn pair-wise distances $\delta_{ij}$ to resolve the highly complex matter of aesthetics in a high-dimensional space.

## 5. Learning Aesthetics

As the visual quality of images is naturally hard to encode in a single scalar and it is hard to match images to discrete bins of aesthetic levels, we aim for directly learning an encoding of a given image in a high-dimensional feature space resembling visual aesthetics in contrast to 1-dim ranking as in [21] (Fig. 6). We will refer to the feature space as the *aesthetic space*. Ranking approaches like [21] predict scalars and inherently assume that image orders are possible on a 1-dim discrete or continuous rating scale. Hence, while a latent group of images might be globally miss-placed in the aesthetic space, our formulation allows to still order the images within the specific group correctly.

### 5.1. Encoding Aesthetics

Inspired by metric learning [5, 13], our approach is to directly optimize *relative* distances

$$\delta : \mathcal{I} \times \mathcal{I} \to \mathbb{R}, \quad (i, j) \mapsto \|\Phi_i - \Phi_j\|_2$$

between encodings $\Phi_i, \Phi_j$ from image pairs $(i, j)$. We use a CNN to learn these encodings, which will be described later in detail. Importantly, this training procedure can be done without associating images to any specifically requested ratings or score from human annotators. Instead, it solely uses the information if two images are similarly aesthetic or not on an almost *arbitrary* scale. We minimize the triplet loss function [13]

$$L_e(a, p, n) = \left[ m + \|\Phi_a - \Phi_p\|_2^2 - \|\Phi_a - \Phi_n\|_2^2 \right]_+ \quad (2)$$

for images $a, p, n$ and some margin $m$. Here, $[x]_+$ denotes the non-negative part of $x$ like the ReLU activation function. This loss resembles a visual comparison, i.e., the distance between two mediocre images $a, p$ should be smaller than the distance to a well-crafted image $n$ and vice versa. Note, our objective function is not directly built on predicting $S(\cdot)$ for a particular image on a *specific* scale and range. To decide whether two images are aesthetically similar or not we use our score $S(i)$ to guide the sampling of the training data consisting of image triplets

$$D = \left\{ (a, p, n) \text{ with } \alpha < \frac{|S(a) - S(p)|}{|S(\bullet) - S(n)|} < \beta \right\} \quad (3)$$

with $a, p \in \bullet$ and $\alpha, \beta \in \mathbb{R}$. Thus, any pair $(a, p)$ with a rather small difference in the score allows for adaptively sampling of much harder negatives $n$ by rejecting triplets with too large differences. An example of such image triplets is shown in Fig. 7. We allow $(a, p)$ to contain images with higher or lower score than $n$ for generating balance training data. This approach has the following advantages:

1. Every single image can be considered during the training *relatively* to other images, which also allows to train on highly debatable images.
2. There is no need to either learn a scalar or solve a binary classification problem in the fashion of ranking [21] or aesthetic-label prediction [35]. Instead, we learn the encoding itself.

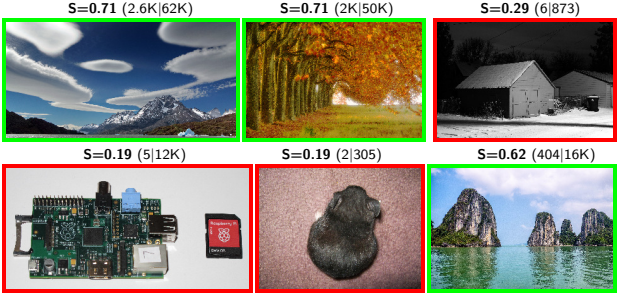

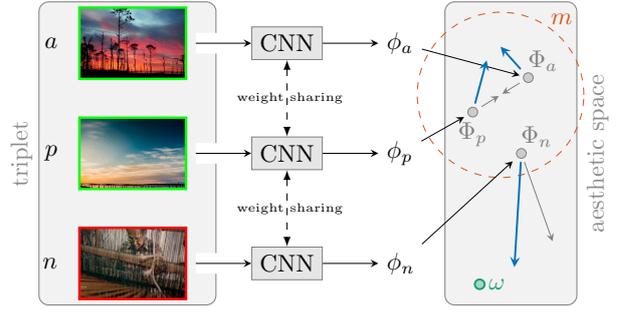

Figure 7. Image triplets $(a, p, n)$ for training with scores $S(i)$. Each triplet consists of either 2 good and 1 bad image concerning its approximated quality or 1 good and 2 bad ones.

Figure 8. Triplet-Loss. For each triplet $(a, p, n)$ with anchor point, we aim at encoding aesthetically similar images $a, p$ nearby and force a larger distance to aesthetic dissimilar images $n$. Adding $L_d$ to $L_e$ alters the update directions wrt. the aesthetic space origin $\omega$.

### 5.2. Rating Aesthetics

As the encodings space $\subseteq \mathbb{R}^d$ is only a partially ordered set, for any two images $x, y$ knowing the aesthetic distance $\|\Phi_x - \Phi_y\|$ has no information if $x$ should be considered as more visually pleasing than $y$. Thus, ordering multiple images is not possible. If an "universally accepted" worst image $\omega$ would exist, then one might simply use the learned distance $\delta(x, \omega)$. But as we are allowed to rotate the entire space, a more practical solution is to force the encoding into a particular direction. We therefore add

$$L_d(a, n) = \text{sign}(s(n) - s(a)) \cdot [\|\Phi_a\| - \|\Phi_n\| + \tilde{m}]_+ \quad (4)$$

as a directional term to the loss function. This leads the triplet loss by reducing the norms of encodings belonging to less visual pleasing images and increases the norms of well crafted images. Note, that we again do not directly use any absolute score values from our data model. Altogether, we minimize the "directional triplet loss":

$$L(a, p, n) = L_e(a, p, n) + L_d(a, n)$$

to get a natural ordering by the Euclidean norm and relative distances. The effect of $L_d$ is depictured in Fig. 8.

### 5.3. Learning the Aesthetic Space

**Network Architecture.** We use the standard ResNet-50 architecture [12] $f_\theta$ with trainable parameters $\theta$ to learn the encodings $\Phi_i = f_\theta(i)$. We add a projection from the *pool5* layer creating a 1000-dimensional descriptor $\Phi$ for each frame. Please refer to [12] for more details. Training was done on two Nvidia Titan X GPUs using stochastic gradient descent with initial learning rate $10^{-3}$ which is divided by 10 when the error plateaus.

**Sampling Training Data.** We randomly sample images from our entire collection on-the-fly according to $D$ in (3). We estimate the cardinality of $D$ as $|D| = 7 \cdot 10^{12}$ from tracking the reject-rate during training. Hence, no data-augmentation is required, which would further influence aesthetics. As ResNet expects the input to have the size $224 \times 224 \times 3$, we resize the original image to match the input dimensions. Although, this down-sampling might remove small details, it keeps the relations of the image content. Further, we are interested in the aesthetics quality, rather than the photo quality from a computational photography viewpoint.

### 5.4. From Space to Scale

To allow for multiple applications, e.g., ranking a set of images, it can be necessary to map our derived encodings within our high-dimensional space to a relative scale. As described earlier, while a latent group of images might be globally miss-placed in the aesthetic space, our formulation allows to still order the images within the specific group correctly. Thus, we simply consider the norm of the encoding $\|\Phi_i\|_2$ as the projection score. Thereby, independently of the positions of the encodings in space, the relations between them stay maintained on the scale.

Table 3. Performance comparison on AVA data set. Different models (top row) with according accuracy (bottom row). Our approach outperforms all models that do not use additional information and even most methods that include additional information during training.

| Additional information during training | | | No additional information | | | | | | | |
|---|---|---|---|---|---|---|---|---|---|---|
| RDCNN [28] | Reg-Rank+Att [21] | Reg-Rank+Att+Cont [21] | Alexnet-FTune [29] | Murray [35] | Reg-Rank [21] | Reg [21] | SPP [28] | DCNN [28] | DMA [29] | **Ours** |
| 74.46 % | 75.48 % | 77.33 % | 59.09 % | 68.00 % | 71.50 % | 72.04 % | 72.85 % | 73.25 % | 74.46 % | **75.83 %** |

## 6. Experimental Results

We pursue two ways of evaluation in quantitative evaluation on the common benchmark set and qualitative evaluation to analyze the internal network mechanism. Further results in combination with applications are presented in Sec. 7 and the supplemental material.

**Quantitative Evaluation.** For a fair comparison to previous approaches, we fine-tune our model network to the distributions of the ratings in the AVA dataset [35]. This is done using a subset of the AVA training data to predict discrete labels instead of relative embeddings. Table 3 shows such a quantitative comparison in accuracy to previous methods. Obviously, using an indirect approach such as ranking (Reg+Rank [21]), which resemble the nature of aesthetic judgments much better than standard approaches like classification [28, 29, 35] yields also better performance on this benchmark set. Ours further boosts this accuracy significantly, which we attribute to the more natural choice of our loss formulation. In contrast to previous work [21, 28], we do not rely on a dedicated neural network architecture using a rather common model design. Results on the left use additional information such as attribute data or content-description. Hence, although we trained on a data set which was constructed with literally no extensive explicit labeling, we outperform *all* previous methods relying solely on ratings they obtained in an expensive process. Further, learning from the consensus of many Flickr users is sufficient to gain higher accuracy (our network) on the AVA benchmark set than recent approaches with additional attributes (Reg+Rank+Attr, RDCNN). Note, these attribute categories are acting essentially as a prior and were selected after consulting professional photographers [21].

We expect to further improve our results when adding more explicit information about the content like in the construction of "Reg+Rank+Att+Cont". As our main focus is to exploit freely available information solely, this explicit meta-information can be image-related comments and tags.

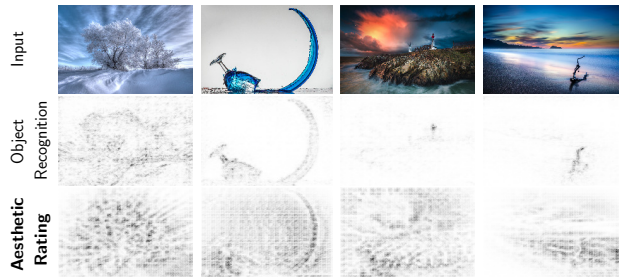

Figure 9. Different photographs (top) and related saliency maps for vanilla ResNet (middle) and our model (bottom) produced by guided-ReLU [39]. Darker region indicates higher influence on the actual network prediction.

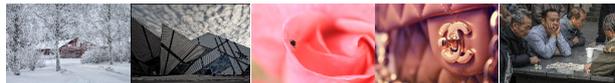

Figure 10. Aesthetically resorted set of photos with decreasing score from our provided tool starting with the visually most pleasing image (left).

**What is the network looking for?** Judging the visual quality of an image is totally different from plain object recognition tasks. When extracting relevant information, which is used by the neural network to perform aesthetics prediction, it is possible to visualize prominent traits in the input. To extract these saliency maps, we use guided-ReLU [39]. It is based on the idea, that large gradients of the output wrt the input have a high impact on the actual network prediction. Fig. 9 highlights those pixels in the input with large impact. Hence, this information is strongly coupled with the encoding in our aesthetic space. It clearly shows how our network considers larger regions in the image space compared to sparse saliency along gradients in the untrained network. More precisely, the network model reveals high synergy effects between surrounding regions in the light-house example in Fig. 9. At same time the vanilla ResNet only focuses on the light-house itself.

## 7. Applications

In order to demonstrate the usability of our approach, we apply our derived aesthetics prediction score to images as well as videos allowing for several applications. Thereby, we map the encodings from space to a relative scale as described in Sec. 5.4 maintaining fine-granular relations.

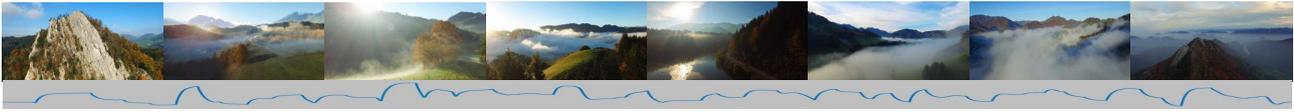

Figure 11. Best video spots. Each frame is extracted at the peaks in the score signal.

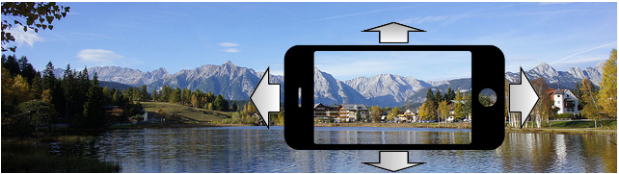

Figure 12. Best handy shot. Based on slight movements in any direction, the application automatically captures the best shot.

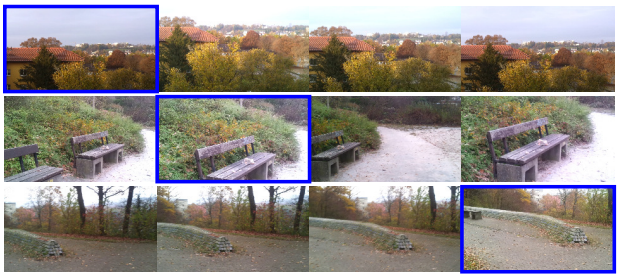

Figure 13. Best predicted image (blue frame) during capturing. The movements were recorded with a mobile device.

**Aesthetic Photo Collection.** First of all, we support resorting an arbitrary photo collection due to our predicted relative aesthetic scores between the images. An example of a small set of aesthetically sorted images is shown in Fig. 10. This tool can facilitate to quickly resort one's holiday collection and directly share the best moments without time-consuming manually browsing of the usually rather large set of pictures.

**Best Handy Shot.** A commonly known situation is that people want to take a picture but are not completely sure what the best shot of the view could be. They tend to take mulitple pictures and just postpone the decision process. This can even lead to missing the one best shot completely. We provide a simple application that allows slightly moving the phone around and temporarily captures a video. The idea is illustrated in Fig. 12. All the single images are then analyzed and rated by our system and the image of the best view is saved. The application supports the user to directly obtain the best aesthetically pleasing image and prevents the time-consuming decision process afterwards. Fig. 13 shows several frames from movements we recorded with a Samsung Galaxy SII phone and the predicted best shots. Sky proportions, saturation and the tension of the overall image layout play an important role within the decision. Due to its small memory footprint of only 102MB containing the network weights, running this application directly on mobile devices is easily possible. Please see the supplemental video for a short demo. This application could further be extended to lead the user to the best shot during the movement while indicating better directions.

**Video Spots.** Similarly, our system is able to find great shots in a video. Those shots can be selected as aesthetic key frames or, e.g., in documentary films, to identify the most wonderful places or spots. Therefore, we calculate a complete prediction curve along the video displaying the aesthetic relation between the frames. Fig. 11 displays an example of a video and the according aesthetic prediction curve. Kalman filtering is applied to smooth the final predictions over time. Extracting the frame scores is done at a speed of 140fps on a NVidia GTX960. Embedding common videos requires only 25% of the actual playback time demonstrating high efficiency and enabling real-time applications. Please see the supplemental material for examples.

## 8. Conclusion

We propose a new data-driven approach which learns to map aesthetics with all its complexity into a high-dimensional feature space. Additionally, we make use of online behavior to incorporate a broad diversity of user reactions as rating aesthetics is a highly subjective task. In detail, we assemble a novel large-scale data set of images from social media content. Hereby, aesthetics ground-truth scores for training are obtained *without* explicitly requesting user ratings in a time-consuming and costly process. Hence, our dataset can be easily extended, as our approach requires effectively no labeling-efforts using freely available information from social media content. The assumption of our underlying model is validated in a user study. To automatically judge aesthetics, we formulate the aesthetic prediction directly as an encoding problem. Consequently, we propose a more naturally loss objective for dealing with the complex task of learning a feature representation of visual aesthetics. Our focus lies on the abstract representation of aesthetics using online media. Thus, we solely rely on a commonly used model architecture and use a much weaker training signal which leads to state-of-the-art results on previous benchmarks. Finally, we confirm the success of our model in several real-world applications, namely, resorting photo collections, capturing the best shot and a smooth aesthetics prediction along a video stream.